\documentclass[12pt,letterpaper]{article}
\usepackage[a4paper, total={7in, 10in}]{geometry}

\usepackage{graphicx}
\usepackage{helvet}
\usepackage{authblk}
\usepackage{hyperref}
\usepackage{amsmath} 
\usepackage{amssymb} 
\usepackage{orcidlink} 
\usepackage{longtable}
\usepackage{caption}
\usepackage[super,comma,sort&compress]  
   {natbib}
\usepackage[right]{lineno} \linenumbers
\usepackage[utf8]{inputenc}
\usepackage[T1]{fontenc}
\usepackage{float}

\makeatletter
\renewcommand{\maketitle}{\bgroup\setlength{\parindent}{0pt}
\begin{flushleft}
  \textbf{\@title}
  
  \@author
\end{flushleft}\egroup}
\makeatother

\title{A feature-stable and explainable machine learning framework for trustworthy decision-making under incomplete clinical data}
\date{}

\author[1\orcidlink{0000-0002-6658-1229}]{Justyna Andrys-Olek}
\author[1\orcidlink{0000-0002-1478-0832}]{Paulina Tworek}
\author[1]{Luca Gherardini}
\author[2]{Mark W. Ruddock}
\author[2]{Mary Jo Kurt}
\author[2]{Peter Fitzgerald}
\author[1, 2\orcidlink{0000-0001-9570-6054}] {Jose Sousa}

\affil[1]{Computational Intelligence, Sano Centre for Computational Personalised Medicine, Krakow, Poland}
\affil[2]{Clinical Studies Group, Randox Laboratories Ltd., Co., Antrim, United Kingdom}
\affil[3]{Coimbra University, Multidisciplinary Institute of Ageing, MIA - Portugal, Coimbra, Portugal}

\affil[*]{Correspondence: j.andrys-olek@sanoscience.org}
\affil[**]{Correspondence: p.tworek@sanoscience.org}
\affil[***]{Correspondence: j.sousa@sanoscience.org}

\begin{document}

\maketitle

\section*{SUMMARY}

Machine learning models are increasingly applied to biomedical data, yet their adoption in high-stakes domains remains limited by poor robustness, limited interpretability, and instability of learned features under realistic data perturbations, such as missingness. In particular, models that achieve high predictive performance may still fail to inspire trust if their key features fluctuate when data completeness changes, undermining reproducibility and downstream decision-making. Here, we present CACTUS (Comprehensive Abstraction and Classification Tool for Uncovering Structures), an explainable machine learning framework explicitly designed to address these challenges in small, heterogeneous, and incomplete clinical datasets.
CACTUS integrates feature abstraction, interpretable classification, and systematic feature stability analysis to quantify how consistently informative features are preserved as data quality degrades. Using a real-world haematuria cohort comprising 568 patients evaluated for bladder cancer, we benchmark CACTUS against widely used machine learning approaches, including random forests and gradient boosting methods, under controlled levels of randomly introduced missing data. We demonstrate that CACTUS achieves competitive or superior predictive performance while maintaining markedly higher stability of top-ranked features as missingness increases, including in sex-stratified analyses.
Our results show that feature stability provides information complementary to conventional performance metrics and is essential for assessing the trustworthiness of machine learning models applied to biomedical data. By explicitly quantifying robustness to missing data and prioritising interpretable, stable features, CACTUS offers a generalizable framework for trustworthy data-driven decision support. Although demonstrated on a bladder cancer cohort, the proposed approach is broadly applicable to other domains where incomplete, high-dimensional data and model transparency are critical.
These findings highlight feature stability as a critical yet underutilised dimension of model evaluation in data-driven biomedical research.

\section*{KEYWORDS}
bladder cancer, biomarkers, decision-making, support system, machine learning, explainable AI, features stability


\section*{INTRODUCTION}

Medical decision-making is a complex process that integrates medical knowledge and experience to formulate a diagnosis or prepare a treatment plan, based on patient health data obtained from tests, medical examinations, and interviews, with the goal of maximising clinical benefit \cite{meddecmak}. 
Despite rapid advances in machine learning for biomedical applications, model evaluation remains dominated by predictive performance metrics, often overlooking whether learned representations remain stable under realistic data perturbations. In practice, biomedical datasets are frequently incomplete, heterogeneous, and subject to acquisition biases, raising concerns about the robustness and reproducibility of model-derived insights.
During diagnosis, most doctors use cognitive shortcuts to formulate a hypothesis by matching an individual's clinical features to typical symptoms of a condition and then confirming it with a series of diagnostic tests. 
A decision made by clinicians must be grounded in the best available evidence, in accordance with an \textit{evidence-based medicine} approach, which requires \textit{application of population-based data to the care of an individual patient (...)} \cite{ebm}. 
It is affected by biases and uncertainties inherent to medical reality, but a clinician must assess the situation and consider all available facts to reduce uncertainty \cite{clindecmak}\cite{meddecmak}. 
Despite extensive training and years of experience, some of these challenges may persist, creating opportunities for frameworks that support decision-making to guide or improve the final evaluation. 
Hence, the search for reliable technological solutions that can serve as a decision aid is well justified. \newline
Applying computer logic in healthcare as a support in human decision-making is not a new concept, as the first serious attempts date back to the 1970s, when the first digital diagnostic assistance based on a decision tree (\textit{INTERNIST-1}) was deployed in Massachusetts General Hospital. 
It was designed by the computer scientist Harry Pople at the University of Pittsburgh to encapsulate the expertise of internist Jack D. Myers and to provide insights for new diagnoses \cite{internist1}. 
The performance of the tool was quickly found to be unsatisfactory due to multiple issues, among which were the attribution of findings to improper causes and the inability to explain its (model) \textit{thinking} \cite{internist1}. 
Years later, many new artificial intelligence (AI) tools remain unreliable or too complex to be trusted for use in medical settings \cite{blackbox,rethinking}. 
One of the main issues is the \textit{black-box problem}, where the internal decision-making processes of AI systems are not transparent or interpretable to users \cite{blackbox}.
For example, the recent boom in Deep Neural Network (DNN) algorithms in healthcare has yielded new insights into how such technology can help. 
However, these models have significant drawbacks: they are highly complex, require substantial computational resources, and demand large datasets for training, which limits their practicality in real-world settings, especially in sensitive areas such as medicine \cite{deepl}.
DNNs comprise multiple layers and employ millions of parameters, such as filters and constraints, making them inherently complex and difficult to interpret \cite{xai}. 
Even the simplest deep models process data in non-trivial ways, making their decision-making process difficult to understand.
As a result, there is a growing trend in the field to prioritise interpretable and explainable machine learning (ML) tools, mostly ones that are intended to be used in high-risk fields like health, banking or criminal justice, where understanding the model’s reasoning is crucial for trust and accountability \cite{stopbbml}.  \newline
One key factor in the adoption of AI-based decision-making support assistants by industry and individuals is trustworthiness. 
In medicine, where the tools clinicians use daily directly affect patients’ health and well-being, establishing trust is even more essential. 
Above all, the model must be explainable and provide clear reasoning behind the decision-making process to both doctors and patients \cite{xai}\cite{trustai}.
A trustworthy AI model must address biases in patient data to ensure fairness while maintaining ethical standards, ensuring patient privacy, demonstrating clinical effectiveness, and delivering reproducible results \cite{aihealth}. \newline
Another obstacle to digital transformation in healthcare is that medical datasets are difficult to use as input for ML models because they often contain substantial amounts of missing and noisy data, making it very challenging to collect complete information for every case \cite{missdat}. 
There are several reasons for missing data, including incomplete medical records, incomplete surveys, and data loss due to lack of patient follow-up.
They can lead to bias, loss of valuable information, reduced statistical power, and generalisability of the findings \cite{missdat2}. 
That is why any solution designed to analyse medical records should be robust to missing data and capable of handling it effectively. \newline
From a data science perspective, instability in feature importance with varying data completeness undermines both interpretability and trust, particularly in high-stakes domains. Yet, systematic evaluation of feature stability remains rare in applied machine learning studies. Addressing this gap requires frameworks that explicitly quantify robustness to missingness while remaining interpretable and data-efficient.
In this paper, we describe how a new ML framework, CACTUS (Comprehensive Abstraction and Classification Tool for Uncovering Structures), designed to handle unbalanced, incomplete, and biased datasets, addresses the aforementioned challenges, positioning it as a trustworthy AI-based framework \cite{cactus}. 
Additionally, we show that CACTUS maintains feature stability, even under conditions of randomly introduced missing data (MCAR). 
This resilience ensures that feature importance remains consistent, reinforcing trust and enhancing the explainability of predictions, which are critical requirements in healthcare and other high-stakes applications \cite{stability}. \newline
The Haematuria Biomarker (HaBio) cohort data were used to classify patients into two groups: those with bladder cancer (BC) and those without bladder cancer (non-BC) \cite{habio}. The goal was to measure how well CACTUS classifies patients based on the full dataset and identify biomarkers specific to patients with BC for the whole population (both genders) and depending on sex (females, males). It was also investigated how introducing an increasing percentage of missing values influences classification and features (biomarkers) stability, as entering missing values simulates a real-world medical scenario, when not all patients underwent the same diagnostic tests or can arise from the loss of a sample batch. Different percentages of missing values (10\%, 20\%, and 30\%) were randomly introduced into each dataset, assuming that missingness affects any feature with the same probability and does not follow any specific pattern. 
In contrast to accuracy-centric evaluation, we argue that stability-aware analysis is essential for trustworthy pattern discovery in real-world data.
\section*{RESULTS AND DISCUSSION}

\subsection*{Feature Stability Metrics as a Foundation for Trustworthy AI in Classifying BC and non-BC Patients}

Beyond predictive performance, we evaluate model behaviour through the lens of feature stability, assessing whether the most informative features remain consistent as data completeness decreases.

Features stability was previously introduced by K. Capała, P. Tworek and J. Sousa (\cite{stability}), to assess how stable the features are when ranked by different ML models. In general, it is desirable for trustworthy AI classification frameorks for medical diagnostics that the most important features for classification remain stable across datasets as the number of randomly introduced missing values increases.

The stability of a given feature can be measured either by directly inspecting its significance value across datasets or by conducting a broader analysis using comparison-based metrics. 
Here, we considered the average relative change in feature importance for the 10 most important features for classification across all used datasets (differing in the number of missing values) for each tested model.
The equation, firstly introduced by Capała et al. (\cite{stability}), is reported in Equation 7 in the supplementary materials.

The smaller the average relative change, the more stable the feature produced by a given model, as it indicates that feature importance remains mostly unaffected with a gradual transition from a complete dataset to its version with 30\% of values removed. The standard deviation is used to quantify how widely feature importance varies across datasets with varying amounts of missing data. 
Hence, low average change and low standard deviation are the ideal characteristics of models that are resilient to missing data.

\begin{figure}[H]
    \centering
    \includegraphics[width=0.6\textwidth]{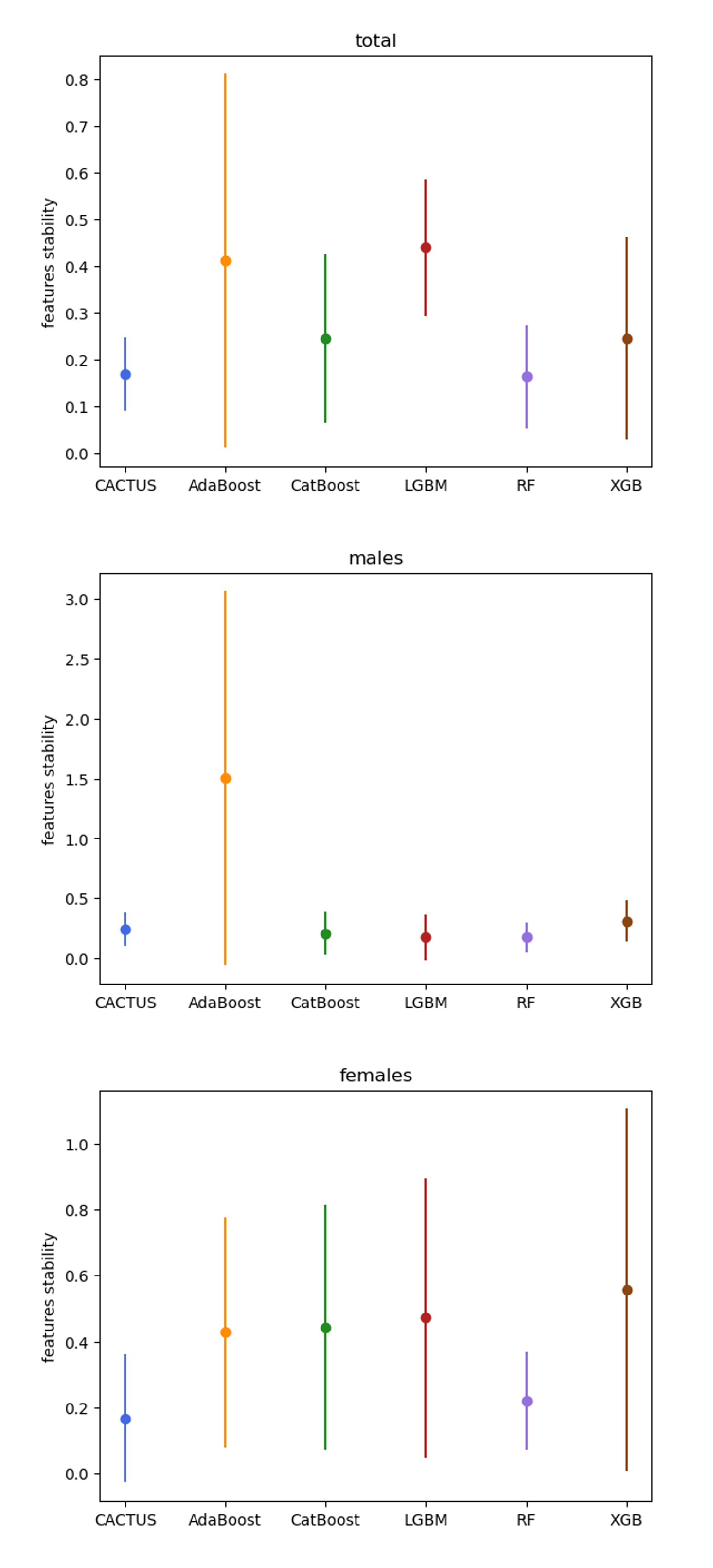}
    \caption{
    \textbf{Stability of features across ML methods}.
    \newline{Average relative change in feature importance calculated for the 10 most important features for classification obtained by each method for: the total dataset (top), the male subjects in the dataset (middle), and the female subjects in the dataset (bottom). 
    \label{fig:stability}
    }}
\end{figure}

Figure \ref{fig:stability} shows that CACTUS achieves low average change in feature importance values in general. For the male and female data subsets, it shows clear superiority over classical ML methods, while for the entire dataset (total), only Random Forest performs better. Furthermore, CACTUS exhibits one of the smallest standard deviation errors, confirming the consistency of the results. 

\subsection*{Another Way of Feature Stability Analysis in Classifying BC and non-BC Patients: The Overlap of Top 10 Ranked Features}

\begin{figure}[H]
    \centering
    \includegraphics[width=0.54\textwidth]{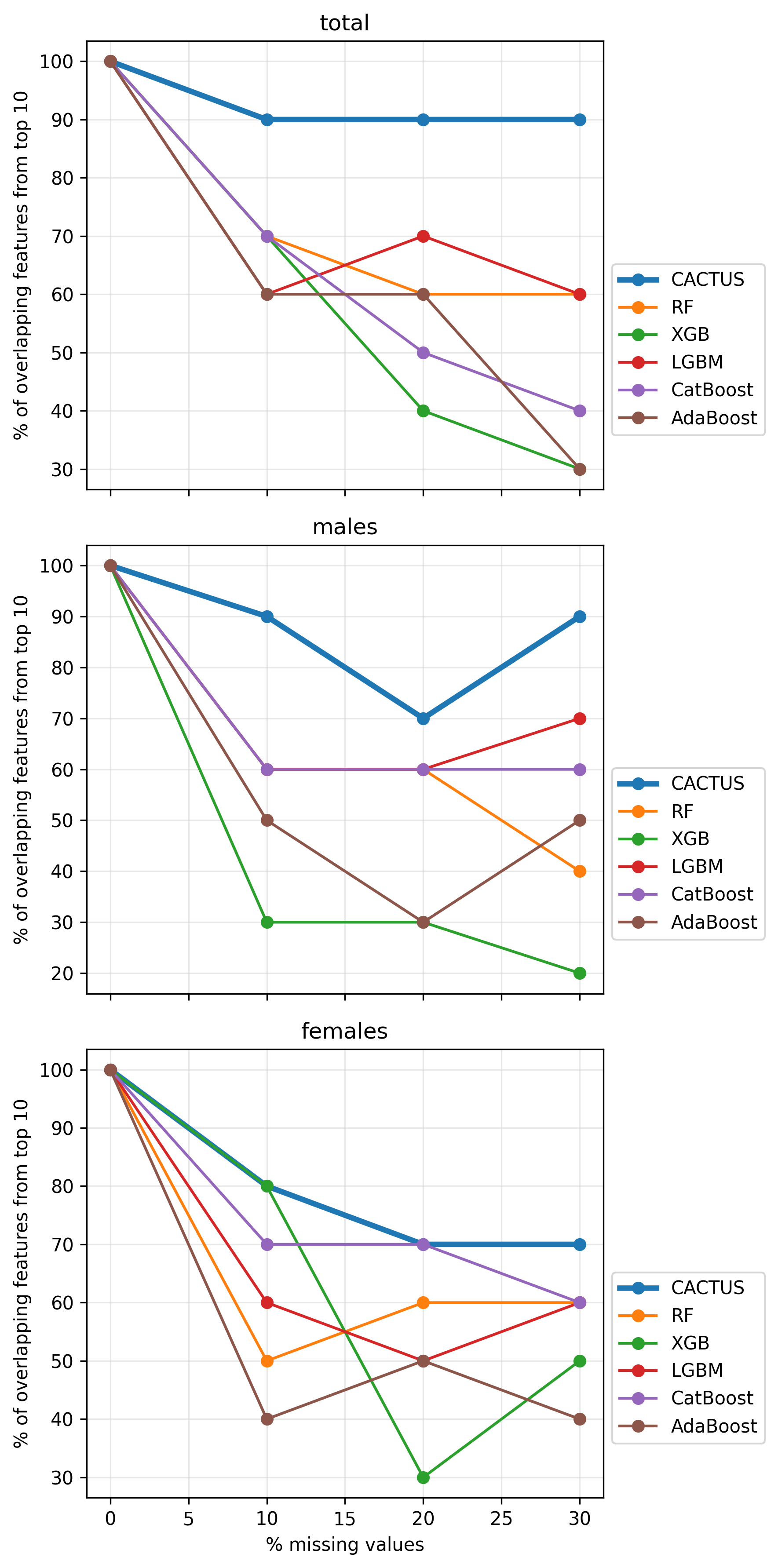}
    \caption{
    \textbf{Overlapping features}.
     \newline{Graphs plotted for each subset from the total population, males and females, presenting percentage of overlapping features from the top 10  most important features for classification with increasing number of missing values in the dataset. 
    \label{fig:overlap}
    }}
\end{figure}

Feature overlap among top-ranked variables provides an intuitive, practitioner-oriented measure of robustness, complementing numerical stability metrics by revealing whether models rely on consistent signals under data perturbations.
The percentage of overlapping features among the 10 most important ones for classification across datasets with different proportions of missing values is another measure of feature stability. 
It is desirable that the top \textit{n} features are stable. 
Observing the same features consistently appearing in the top 10 ranks for each dataset variant (complete and with increasing percentage of removed values up to 30\%) can provide confidence to healthcare professionals that these features are indeed important in the classification, and consequently in the diagnostic process. 
This aspect is particularly valuable from the medical point of view, as a clinician might want to focus primarily on the \textit{n} top-ranked features without inspecting the stability values in detail. That way, an ML model can provide direct support in a decision-making process. Although individual features may not always remain stable (Fig. \ref{fig:stability}), CACTUS gives more consistent and repetitive results across all three datasets (total, males, and females) when considered as a group, as illustrated in Figure \ref{fig:overlap}.  Classical methods cannot produce such stability for all three subsets. RF and CatBoost are the second best after CACTUS only for the total population and the male datasets, but CatBoost fails for females, where the second best is RF and LGBM. Features for CACTUS and LGBM are additionally shown in tables \ref{tab:features-tot-cactus}-\ref{tab:features-females-rf}, while results for the rest of the tested methods are available in Supplementary Information (Tab. S1-S12).

\renewcommand{\arraystretch}{1.1}
\begin{table}[H]
\centering
\captionsetup{  labelfont=bf,
  textfont=bf,
  font=scriptsize}
\caption{The most important 10 features for classification obtained through CACTUS analysis for the combined dataset (total population). The prefix \textit{(s)} indicates that the biomarker is measured in blood rather than in urine (default).}
\label{tab:features-tot-cactus}
\scriptsize
\scshape
    \begin{tabular}{|c|c|c|c|c|c|c|c|c|}
    \hline
     rank & complete & rank  & complete +10\% & rank & complete +20\% & rank & complete +30\%& rank \\
    number &   & value & missing values & value & missing values & value &  missing values & value \\
    \hline
      1 & haematuria & 0.304 & clusterin & 0.317 & clusterin & 0.253 & clusterin & 0.220 \\
    \hline
    2 & clusterin & 0.297 & nse & 0.295 & fas & 0.219 & cystatin-b & 0.220 \\
    \hline
    3 & nse & 0.279 & haematuria & 0.257 & tar expos. & 0.215 & nse & 0.210 \\
    \hline
    4 & {tar expos. }& 0.252 & microalbumin & 0.248 & nse & 0.214 & vegf & 0.208 \\
    \hline
    5 & il-\tiny{1}\scriptsize{$\alpha$} & 0.242 & (s)pai-\tiny{1}\scriptsize{/tpa} & 0.241 & (s)pai-\tiny{1}\scriptsize{/tpa} & 0.213 & haematuria & 0.206 \\ 
    \hline 
    6 & microalbumin & 0.242 & cystatin-b & 0.230 & microalbumin & 0.199 &  il-\tiny{1}\scriptsize{$\alpha$} & 0.197 \\
    \hline
    7 & (s)pai-\tiny{1}\scriptsize{/tpa} & 0.239 &  il-\tiny{1}\scriptsize{$\alpha$} & 0.229 & haematuria & 0.186 & fas & 0.190 \\
    \hline 
    8 & fas & 0.230 & bta & 0.221 & tpa & 0.186 & (s)pai-\tiny{1}\scriptsize{/tpa} & 0.190 \\
    \hline 
    9 & bta &0.229 & fas& 0.219& cystatin-b&0.185 & {tar expos. }&0.187 \\ 
    \hline
    10 &cystatin-b & 0.229& tpa& 0.218& il-\tiny{8}& 0.172& bta& 0.180\\ 
    \hline
    \end{tabular}  
\bigskip
\renewcommand{\arraystretch}{1.1}
\centering
\scriptsize
\captionsetup{font=scriptsize}
\caption{The most important 10 features for classification obtained through LGBM analysis for the combined dataset (total population). The prefix \textit{(s)} indicates that the biomarker is measured in blood rather than in urine (default).}
\label{tab:features-tot-rf}
\scshape
    \begin{tabular}{|c|c|c|c|c|c|c|c|c|}
    \hline
    rank & complete & rank  & complete +10\% & rank & complete +20\% & rank & complete +30\%& rank \\
    number &   & value & missing values & value & missing values & value &  missing values & value \\
    \hline
     1 & nse & 390.4 & nse & 309.3 & nse & 251.5 & psa/tpsa & 214.9\\
    \hline
    2  & psa/tpsa & 192.0 & psa/tpsa & 260.5 & {tar expos. } & 155.2 &  yrs of smoking&172.9 \\
    \hline
    3  &yrs of smoking  & 178.7 & (s)pai-\tiny{1}\scriptsize{/tpa}  & 169.8 & clusterin  & 140.8 & (s)pai-\tiny{1}\scriptsize{/tpa} & 171.4\\
    \hline
    4  & haematuria & 169.2 & haematuria & 163.4 & haematuria & 115.7 & nse & 152.6 \\
    \hline
    5  & (s)egf & 123.1 &(s)egf  & 100.4 & perk &108.3  &  midkine&129.7 \\
    \hline 
    6  & (s)pai-\tiny{1}\scriptsize{/tpa}  & 100.5 & il-\tiny{18} & 99.7 & mmp\tiny{9} & 105.8 & haematuria & 121.7 \\
    \hline
    7  & recurrent uti & 92.2 & prolactin & 90.6 &  psa/tpsa & 101.6 &   il-\tiny{1}\scriptsize{$\alpha$}&117.6 \\
    \hline 
    8  &egf  & 79.6 &acr  &  88.1& (s)egf & 98.6 & (s)egf &104.1 \\
    \hline 
    9  & clusterin & 77.9 & clusterin & 87.6 & yrs of smoking &88.7  & cystatin-b & 76.6\\
    \hline
    10  & d-dimer & 76.1 & fas  & 77.5 & (s)pai-\tiny{1}\scriptsize{/tpa} &75.6  &acr  & 67.4\\
    \hline
    \end{tabular}
\bigskip
\renewcommand{\arraystretch}{1.1}
\centering
\scriptsize
\captionsetup{font=scriptsize}
\caption{The most important 10 features for classification obtained through CACTUS analysis for the males dataset. The prefix \textit{(s)} indicates that the biomarker is measured in blood rather than in urine (default).}
\label{tab:features-males-cactus}
\scshape
    \begin{tabular}{|c|c|c|c|c|c|c|c|c|}
    \hline
    rank & complete & rank  & complete +10\% & rank & complete +20\% & rank & complete +30\%& rank \\
    number &   & value & missing values & value & missing values & value &  missing values & value \\
    \hline
     1 & nse & 0.298 & nse & 0.281 & fas & 0.224 & (s)psa/tpsa & 0.207 \\
    \hline
    2 & {tar expos. } & 0.281 & vegf & 0.260 & nse & 0.223 & {tar expos. } & 0.202  \\
    \hline
    3 & clusterin & 0.273 & clusterin & 0.245 & clusterin & 0.216 & nse & 0.198 \\
    \hline
    4 & fas & 0.266 & fas & 0.244 & prolactin & 0.213 & fas & 0.185 \\
    \hline
    5 & prolactin & 0.262 & cystatin-b & 0.231 & cystatin-b & 0.212 & prolactin & 0.181 \\ 
    \hline 
    6 & vegf & 0.258 & {tar expos. } & 0.227 & il-\tiny{1}\scriptsize{$\alpha$} & 0.202 &  vegf & 0.174 \\
    \hline
    7 & cystatin-b & 0.253 &  prolactin &0.223 & gro & 0.194 & midkine & 0.171 \\
    \hline 
    8 & midkine &0.237 & (s)psa/tpsa & 0.220 & {tar expos. } & 0.194 & cystatin-b & 0.170 \\
    \hline 
    9 & (s)psa/tpsa &0.227 & tpa& 0.210& vegf&0.188 & il-\tiny{7}&0.166 \\ 
    \hline
    10 &tpa & 0.223&  il-\tiny{1}\scriptsize{$\alpha$}& 0.1993& cxcl\tiny{16}& 0.186& tpa& 0.165\\ 
    \hline
    \end{tabular}
\end{table}

\begin{table}[H]

\clearpage
\renewcommand{\arraystretch}{1.1}
\centering
\scriptsize
\captionsetup{  labelfont=bf,
  textfont=bf,
  font=scriptsize}
\caption{The most important 10 features for classification obtained through LGBM analysis for the male dataset. The prefix \textit{(s)} indicates that the biomarker is measured in blood rather than in urine (default).}
\label{tab:features-males-rf}
\scshape
    \begin{tabular}{|c|c|c|c|c|c|c|c|c|}
    \hline
    \textbf{rank} & \textbf{complete} & \textbf{rank}  & \textbf{complete +10\%} & \textbf{rank} & \textbf{complete +20\%} & \textbf{rank} & \textbf{complete +30\%} & \textbf{rank} \\
    \textbf{number} &   & \textbf{value} & \textbf{missing values} & \textbf{value} & \textbf{missing values} & \textbf{value} &  \textbf{missing values} & \textbf{value} \\
    \hline
    1 & nse  & 318.2 & nse &228.3 &nse  &166.7 &nse  &207.9\\
    \hline
    2 &yrs of smoking &  172.5 & psa/tpsa & 151.2 &psa/tpsa &  127.4& psa/tpsa& 136.2 \\
    \hline
    3 & prolactin  & 127.3 & tnf-{$\alpha$} &115.2 & yrs of smoking &102.5 & prolactin &119.9\\
    \hline
    4 & psa/tpsa  & 123.3 & prolactin & 114.7& clusterin &91.8 &{tar expos. }& 111.6  \\
    \hline
    5 & egf  & 98.3& (s)pai-\tiny{1}\scriptsize{/tpa} & 107.4& il-\tiny{8} & 88.9& tgf-{$\beta$}\tiny{1} &89.8\\
    \hline 
    6 &  ngal & 96.6 & haematuria  &103.3 &  tgf-{$\beta$}\tiny{1}&80.9 & fas &84.1\\
    \hline
    7 &  (s)pai-\tiny{1}\scriptsize{/tpa}  & 85.8 & yrs of smoking & 99.8& prolactin & 78.9& haematuria &82.0\\
    \hline 
    8 &  haematuria & 81.8 & (s)crp & 88.5& (s)pai-\tiny{1}\scriptsize{/tpa}  & 73.4& yrs of smoking &76.8\\
    \hline 
    9 & (s)egf  &68.4  &  tgf-{$\beta$}\tiny{1} &82.0 & haematuria & 70.9 &  (s)egf& 75.1\\
    \hline
    10 & lasp-\tiny{1}  &67.1  & crp & 73.3& (s)vegf & 64.4& ngal &67.9\\
    \hline
    \end{tabular}

\bigskip
\renewcommand{\arraystretch}{1.1}
\centering
\scriptsize
\captionsetup{font=scriptsize}
\caption{The most important 10 features for classification obtained through CACTUS analysis for the females dataset. The prefix \textit{(s)} indicates that the biomarker is measured in blood rather than in urine (default).}
\label{tab:features-females-cactus}
\scshape
    \begin{tabular}{|c|c|c|c|c|c|c|c|c|}
    \hline
     \textbf{rank} & \textbf{complete} & \textbf{rank}  & \textbf{complete +10\%} & \textbf{rank} & \textbf{complete +20\%} & \textbf{rank} & \textbf{complete +30\%} & \textbf{rank} \\
    \textbf{number} &   & \textbf{value} & \textbf{missing values} & \textbf{value} & \textbf{missing values} & \textbf{value} &  \textbf{missing values} & \textbf{value} \\
    \hline
    1 &haematuria& 0.555 & haematuria & 0.497& haematuria & 0.512 & microalbumin &0.364 \\ 
    \hline
    2 & microalbumin&  0.437& microalbumin & 0.453& clusterin & 0.429 & haematuria&0.356 \\ 
    \hline
    3 &il-\tiny{1}\scriptsize{$\alpha$} & 0.418 & bta &0.417 & il-\tiny{1}\scriptsize{$\alpha$} & 0.422 & il-\tiny{8} & 0.328\\ 
    \hline
    4 & clusterin&0.418  & clusterin &0.414& microalbumin & 0.421 & acr& 0.310\\ 
    \hline
    5 & bta &  0.404& il-\tiny{1}\scriptsize{$\alpha$} &0.408 & cxcl\tiny{16}  &0.401  & bta &0.304 \\ 
    \hline 
    6 &mcp-\tiny{1} & 0.401 & il-\tiny{8} &0.406 & bta & 0.392 &d-dimer &0.300 \\ 
    \hline
    7 & il-\tiny{8}&0.399  & mcp-\tiny{1}  &0.381 & mcp-\tiny{1} & 0.361 &il-\tiny{7} & 0.294\\ 
    \hline 
    8 & vegf & 0.393 &  protein &0.381 & acr &  0.351& il-\tiny{1}\scriptsize{$\alpha$}& 0.287\\ 
    \hline 
    9 &il-\tiny{13} &0.387& nse &0.376 &   il-\tiny{8}& 0.332 &il-\tiny{13} & 0.285\\ 
    \hline
    10 & (s)pai-\tiny{1}\scriptsize{/tpa} & 0.385 & (s)pai-\tiny{1}\scriptsize{/tpa} & 0.365& d-dimer & 0.329 & clusterin &0.283 \\ 
    \hline
    \end{tabular}
\bigskip
\renewcommand{\arraystretch}{1.1}
\centering
\scriptsize
\captionsetup{font=scriptsize}
\caption{The most important 10 features for classification obtained through LGBM analysis for the females dataset. The prefix \textit{(s)} indicates that the biomarker is measured in blood rather than in urine (default).}
\label{tab:features-females-rf}
\scshape
    \begin{tabular}{|c|c|c|c|c|c|c|c|c|}
    \hline
     \textbf{rank} & \textbf{complete} & \textbf{rank}  & \textbf{complete +10\%} & \textbf{rank} & \textbf{complete +20\%} & \textbf{rank} & \textbf{complete +30\%} & \textbf{rank} \\
    \textbf{number} &   & \textbf{value} & \textbf{missing values} & \textbf{value} & \textbf{missing values} & \textbf{value} &  \textbf{missing values} & \textbf{value} \\
    \hline
    1 & haematuria  & 122.8 & haematuria &85.9 & haematuria &159.9 & haematuria &80.0\\
    \hline
    2 & microalbumin  & 80.5 & microalbumin &75.3 & clusterin & 92.3&microalbumin  &65.7\\
    \hline
    3 & il-\tiny{13}  &  48.5& il-\tiny{1}\scriptsize{$\alpha$} &67.7 &il-\tiny{1}\scriptsize{$\alpha$}  &61.9 & mcp-\tiny{1} &43.4\\
    \hline
    4 &  (s)pai-\tiny{1}\scriptsize{/tpa}  & 37.9 &yrs of smoking  & 52.3& gro &41.1 & il-\tiny{8} &42.4\\
    \hline
    5 &  clusterin & 35.0 & bta &51.6 & microalbumin &37.9 & clusterin & 21.3\\
    \hline 
    6 & bta  &30.2  & clusterin &41.8 & yrs of smoking & 35.9&  yrs of smoking&20.8\\
    \hline
    7 &   yrs of smoking & 30.1 &  progranulin& 35.6&  (s)cystatin-c&29.8 & bta &17.2\\
    \hline 
    8 &  mcp-\tiny{1} &29.1  & fabp-a & 27.0& acr &24.1 &il-\tiny{12}\scriptsize{p}\tiny{70}  &14.8\\
    \hline 
    9& recurrent uti  &25.1  & nse &25.9 & cxcl\tiny{16} &23.2 & cxcl\tiny{16} &14.4\\
    \hline
    10& triglicerides  & 23.0 &il-\tiny{13}  & 25.5& (s)pai-\tiny{1}\scriptsize{/tpa} &20.0 & cystatin-c &13.0\\ 
    \hline
    \end{tabular}
\end{table}

\subsection*{Performance of ML Methods in BC and non-BC Patients Classification}
The gold standard for ML models evaluation in comparison to the metrics obtained by them. Specific metrics should be selected depending on the problem being solved. In the BC and non-BC patients classification task, we primarily focus on two metrics: balanced accuracy and recall (sensitivity). 
Balanced accuracy is a measure of how frequently an ML model makes correct predictions when dealing with imbalanced datasets, while recall is the ability of the model to spot positive cases, here BC patients. 
Recall (sensitivity) is particularly important in medical applications, where we aim to identify as many positive cases as possible, i.e., people who are sick or at risk of becoming sick (here, BC patients).
Additional metrics: F1 score, accuracy and precision are calculated and presented in Supplementary Materials (Figures S2-S4).

\begin{figure}[h!]
    \centering
    \includegraphics[width=0.6\textwidth]{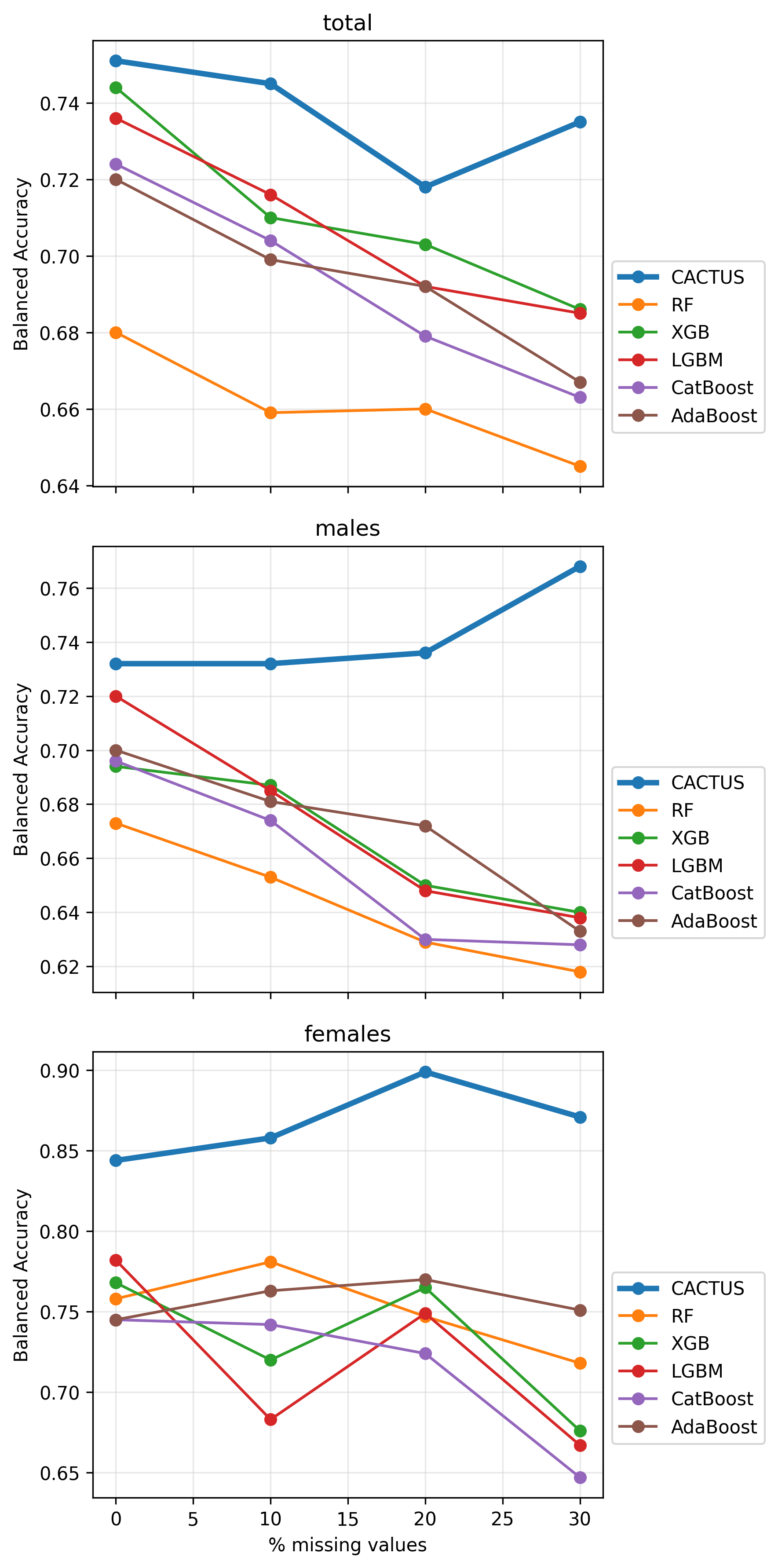}
    \caption{
    \textbf{Balanced Accuracy (BA)}.
     \newline{BA calculated for each subset (total, males and females) with increasing number of missing values in the dataset. 
    \label{fig:ba}
    }}
\end{figure}

\begin{figure}[h!]
    \centering
    \includegraphics[width=0.6\textwidth]{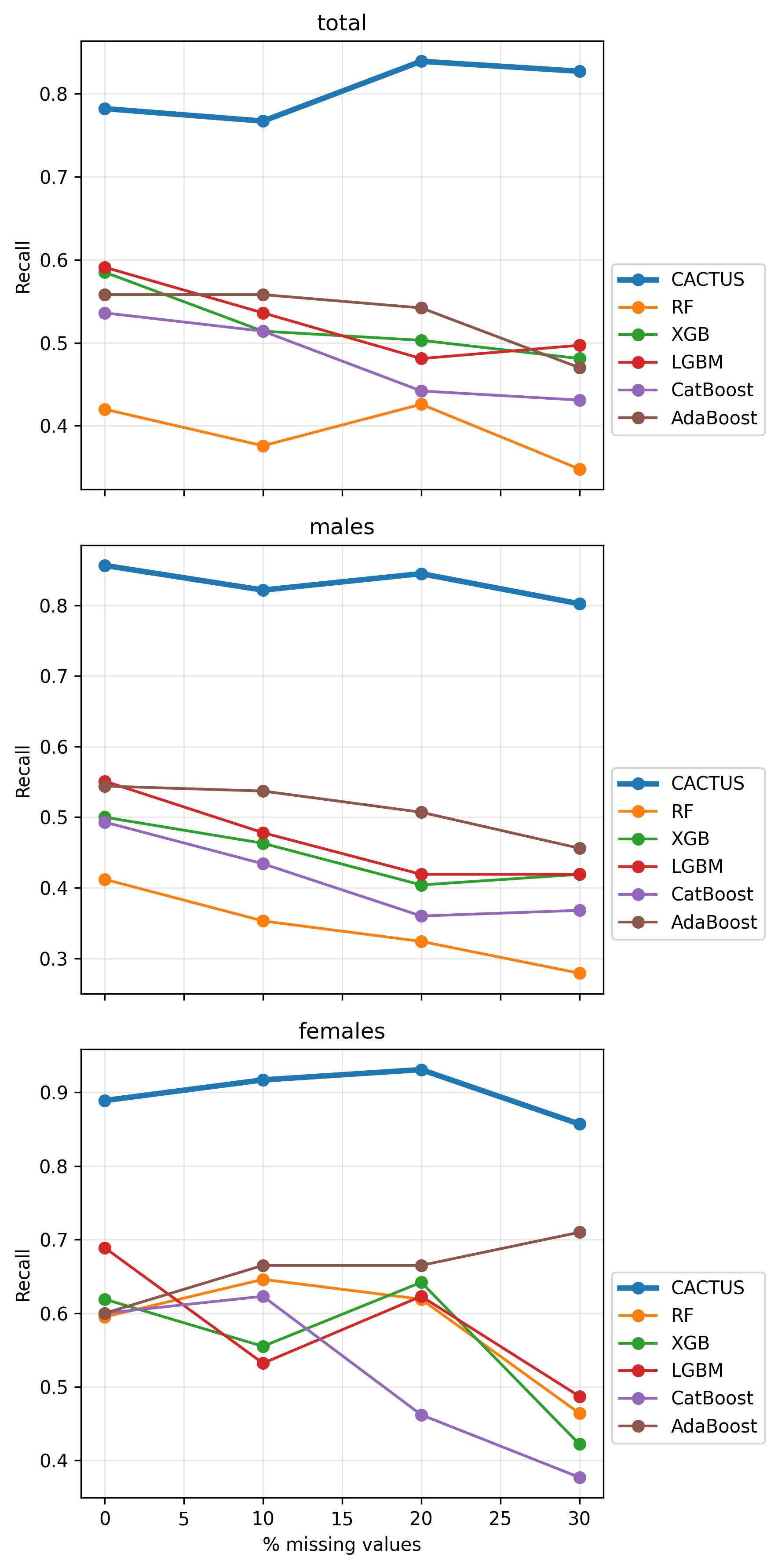}
    \caption{
    \textbf{Recall (sensitivity)}.
     \newline{Sensitivity calculated for each subset (total, males and females) with increasing number of missing values in the dataset. 
    \label{fig:recall}
    }}
\end{figure}

CACTUS produced better results regarding balanced accuracy and recall when compared with other classical ML models (Fig. \ref{fig:ba}-\ref{fig:recall}). CACTUS achieved higher balanced accuracy, especially for females and males subsets (Fig. \ref{fig:ba}, bottom and middle), whereas for recall, it is superior for all three datasets and all levels of introduced missing values (Fig. \ref{fig:recall}). Because the model performance is just one of the metrics available for ML model evaluation, other information, like feature stability, should be included in the analysis. The high stability of the features, together with high performance measured with metrics, shows that CACTUS is effective for the analysis of small and incomplete medical datasets. 
\newline
\clearpage
\clearpage

\subsection*{The 10 most important biomarkers for classifying patients as BC or non-BC patients  for both sexes identified by CACTUS}

As proven above, CACTUS achieves overall better results over classical ML models in the case of differentiating between BC and non-BC patients in all three datasets, with and without stratification by sex. A sex-based comparison of features is conducted, as performance metrics, especially recall, show significant differences between males and females, suggesting sex-specific biomarkers in bladder cancer development. Descriptions of 10 of these biomarkers identified by CACTUS, together with heatmaps illustrating their discriminatory power between BC and non-BC cases depending on sex, are presented below. The analysis of the 10 most important features performed for the total population is provided in the supplementary material (Fig. S1). 

\begin{figure}[h!]
    \centering
    \includegraphics[width=0.7\textwidth]{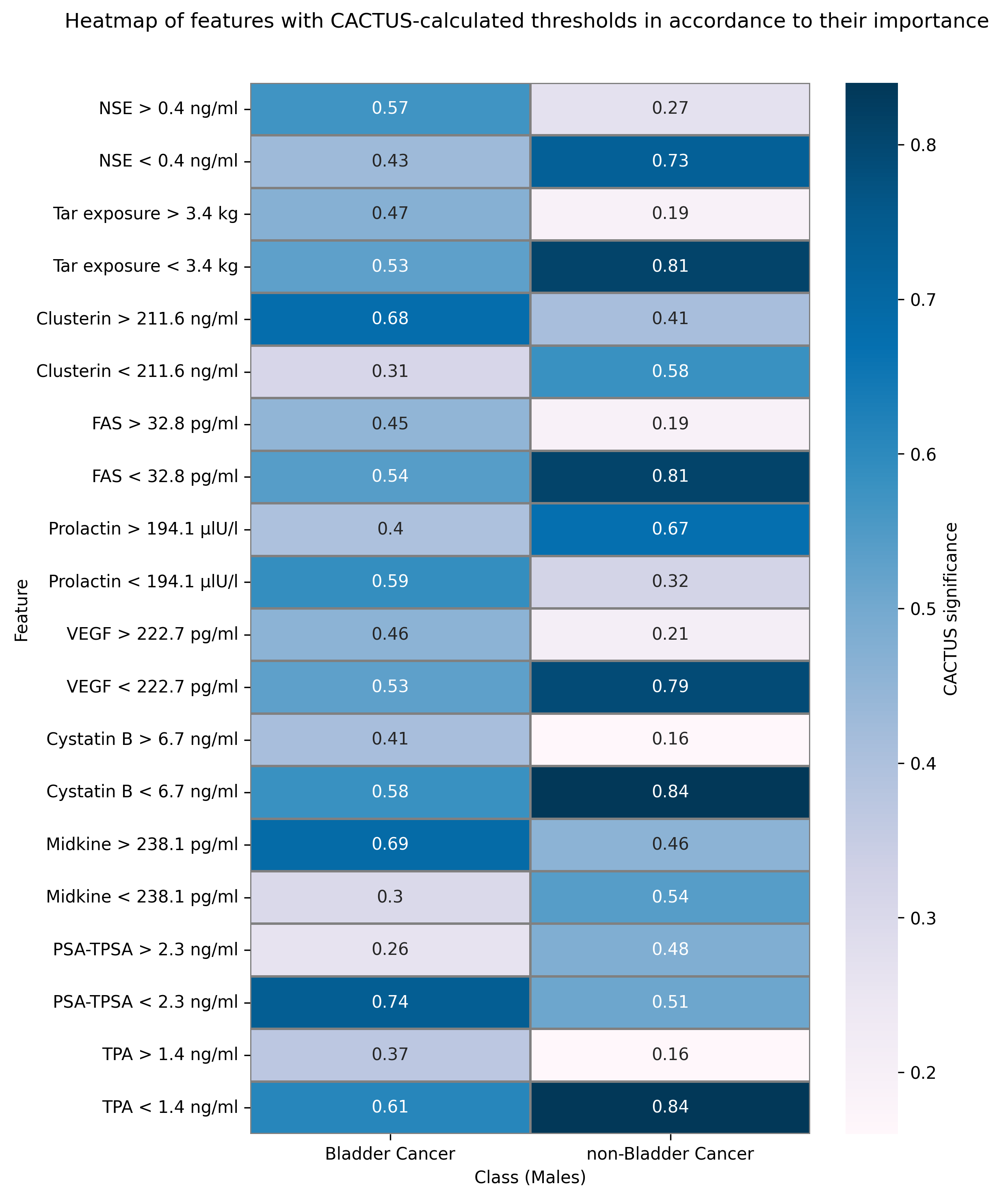}
    \caption{
    \textbf{Heatmap for males subset}.
     \newline{A heatmap showing the 10 most important features for BC/non-BC case classification in the male subset, with thresholds and significance values used to produce the ranks. Note: the thresholds shown on the graph represent the best value that separates the two classes (BC and non-BC cases) and do not correspond to thresholds set by medical institutions for diagnosis.
    \label{fig:males_hm}
    }}
\end{figure}
\begin{enumerate}
  \item \textbf{NSE (urine)}: neuron-specific enolase is a protein specific to neurons and peripheral endocrine cells. It serves as a biomarker for small cell lung carcinoma, as increased body fluid levels of NSE may occur with malignant proliferation and thus can be of value in diagnosis, staging and treatment of related neuroendocrine tumours (NETs)\cite{nse}. However, no link between NSE and bladder cancer has been established to this date. 
  \item \textbf{Tar exposure}:  it refers to the burnt matter left after smoking a cigarette. It forms a sticky layer on the respiratory mucosa, impairing its functions and leading to many lung diseases \cite{tar1}. Various types of cigarettes leave different amounts of tar in the lungs, depending on how they are manufactured \cite{tar1}.  The link between tar exposure and upper airway cancers is well known. In general, smoking is an established risk factor for bladder cancer and tar plays an important role in carcinogenesis \cite{tar2}.
  \item \textbf{Clusterin (urine)}: Clusterin is a protein found widely in blood and tissues. It can exist in different forms and have distinct functions, depending on the location and expression levels. Although sCLU (secreted form of CLU) is known to play crucial roles in lipid transport and cellular lysis and adhesion under physiological conditions, elevated levels are often noted in several pathologies, like Alzheimer’s disease, fibrosis, cardiovascular diseases or cancers \cite{clusterin1}. Increased expression of clusterin is already linked with bladder cancer, making it a promising candidate biomarker for disease detection and prognosis \cite{clusterin2}. 
  \item \textbf{FAS/CD95 (urine)}: FAS cell surface death receptor is a protein that starts the molecular cascade leading to apoptosis upon binding with its ligand (FASL). A soluble form of CD95 (a product of alternative mRNA splicing) is, on the other hand, an antiapoptotic factor. Elevated levels of sFAS are linked with many cancers, including small cell lung, ovarian, endometrial, adrenocortical and colorectal cancers \cite{fas1,fas2}. Moreover, sFAS has been identified as a potential biomarker not only for predicting outcomes in patients with bladder cancer but also for detecting the disease itself \cite{fas3,fas4}. 
  \item \textbf{Prolactin (serum)}: Prolactin is a hormone produced by the pituitary gland, which regulates testosterone production and sexual functions in males. Elevated prolactin levels are observed in both nodular hyperplasia and prostate cancer patients \cite{prolactin}, and our results, shown in Figure \ref{fig:males_hm}, indicate a negative correlation between prolactin and bladder cancer and a positive correlation between prolactin and non-bladder cancer patients. 
  \item \textbf{VEGF (urine)}: vascular endothelial growth factor (VEGF) is a cytokine secreted by osteoblasts, macrophages, cancer cells, megakaryocytes and platelets. It plays an important role in angiogenesis by stimulating vascular permeability and vessel growth \cite{vegf}. It has been proven that an increased level of VEGF in tumours is associated with more intensive cell proliferation and metastasis, which translates to poor prognosis in cancer patients \cite{vegf2,vegf3}. Moreover, it has been recognised as a potential biomarker associated with bladder cancer \cite{vegf4}.  
  \item \textbf{Cystatin B (urine)}: Cystatin B is a member of cystatin proteins which act as inhibitors of cathepsin proteases by binding to them, thus preventing cellular protein degradation, particularly from lysosomal enzymes \cite{cysb}. Increased urinary levels of cystatin B have been observed in patients with bladder cancer and show a strong association with tumour grade and disease stage \cite{cysb2}. 
  \item \textbf{Midkine (urine)}: Midkine, also known as neurite growth-promoting factor 2, is a small protein that takes part in angiogenesis, fibrinolysis, cell division and migration. Under physiological conditions, its activity is restricted to specific tissue types and remains generally low \cite{midkine1}. However, its overexpression has been linked with inflammatory responses and carcinogenesis in at least 20 distinct types of cancer, including bladder \cite{midkine2,midkine3}.
  \item \textbf{PSA/tPSA (serum)}: it refers to the measurement of the ratio of free PSA (prostate-specific serum antigen) to total PSA (tPSA), which enables more precise assessment of prostate cancer risk. Low PSA/tPSA ratio indicates high risk of prostate cancer \cite{}, while a high PSA/tPSA ratio indicates benign changes in the gland. PSA is a biomarker of prostate cancer, but no record of a link between PSA and bladder cancer has been found to date. 
  \item \textbf{tPA}: (tissue plasminogen activator) facilitates the breakdown of clots. The level of free tPA can be an informative biomarker in breast cancer, when it forms a complex with PAI-1 protein \cite{tpa1}. However, there are no studies to date to show the link between standalone free tPA and bladder cancer. 

\end{enumerate}
\clearpage
\begin{figure}[h!]
    \centering
    \includegraphics[width=0.7\textwidth]{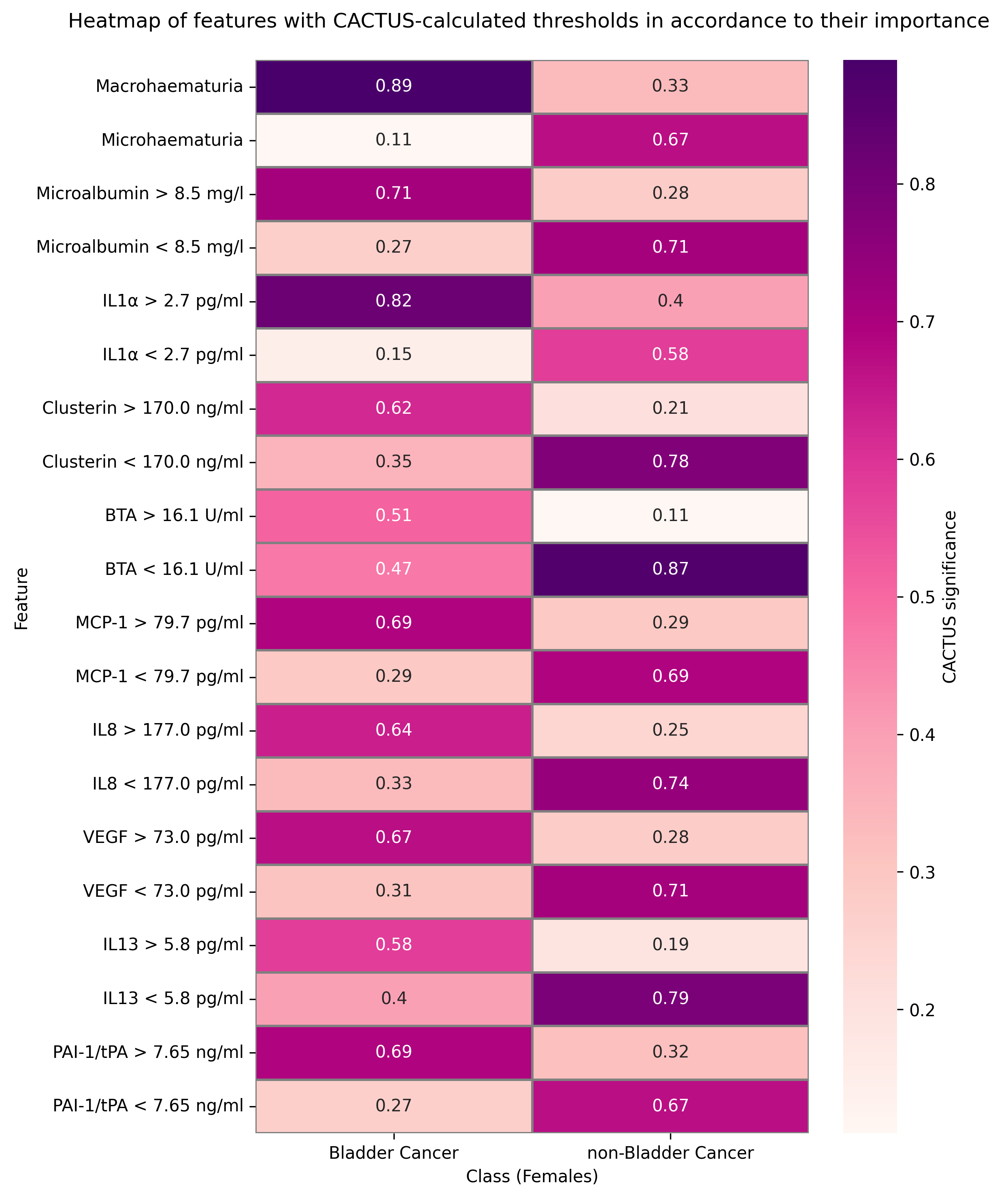}
    \caption{
    \textbf{Heatmap for females subset}.
     \newline{A heatmap showing the 10 most important features for BC/non-BC case classification in the male subset, with thresholds and significance values used to produce the ranks. Note: the thresholds shown on the graph represent the best value that separates the two classes (BC and non-BC cases) and do not correspond to thresholds set by medical institutions for diagnosis.
    \label{fig:females_hm}
    }}
\end{figure}

To avoid repetition, the descriptions of overlapping features with MALES heatmap (Figure \ref{fig:males_hm}), Clusterin (urine) and VEGF (urine), are omitted here.  

\begin{enumerate}
    \item \textbf{Haematuria}: A presence of red blood cells (RBCs) in the urine. When RBCs are visible to the naked eye, the condition is known as macrohaematuria. If they are visible only under a microscope, the condition is called microhaematuria. Haematuria indicates that pathological processes may take place within the urinary tract. These can be benign in nature, e.g. kidney stones, bacterial infection, menstruation (females) or prostate enlargement (males), but may also signal the presence of malignancies like bladder cancer \cite{hemat,hemat2}.
    \item \textbf{Microalbuminuria}: A condition defined by the excretion of 30–300 mg of albumin in urine over a 24-hour period, confirmed in at least two out of three urine samples. It is a crucial indicator of kidney damage, as normally, proteins are retained in the blood by the renal glomeruli and not secreted in the urine. Two cohort studies have established a correlation between higher urinary albumin levels and an increased risk of developing cancers (urinary tract, lung, haematological), independently of kidney function \cite{micro1,micro2}.
    \item\textbf{BTA (urine)}: The bladder tumor antigen (BTA) test is designed to detect human complement factor-H related protein (hCFHrp). It has been found that bladder cancer cell lines can produce and release hCFHrp, whereas normal epithelial cells do not. This suggests that hCFHrp secretion is associated with malignant transformation \cite{bta,bta2}. However, the presence of haematuria in urine samples can interfere with test results, leading to false positives. Hence, blood presence should be considered a confounding factor in diagnostic assessments with use of a BTA test \cite{bta3}.
    \item \textbf{IL-1$\alpha$ (urine)}: Interleukin-1$\alpha$ is a pro-inflammatory protein that promotes fever and sepsis, typically released as a result of cell death or tissue damage \cite{il1a}. It is a key factor in the pathogenesis of multiple inflammation-related conditions, including various types of cancers. In the context of HaBio cohort, it is worth noting that elevated IL-1$\alpha$ levels were associated with malignant progression of bladder cancer \cite{il1a,il1a2}. 
    \item \textbf{MPC-1 (urine)}: MCP-1 stands for monocyte chemoattractant protein 1. It is a small cytokine that plays an important role in immunoregulation and inflammation \cite{mcp1}. It has been shown that MCP-1 is associated with infections, bowel disease, diabetes, and several types of cancer \cite{mcp1_2,mcp1_3}. Furthermore, urinary MCP-1 concentration has been proposed as a potential prognostic indicator of bladder cancer progression \cite{mcp1_3}. 
    \item \textbf{IL-8 (urine)}: Another protein from the interleukin group. IL-8 is a chemokine that plays an important role in the immune response to infection through chemotaxis \cite{il8}. It also promotes cell migration, angiogenesis, and metastasis, and is known as a pro-cancerogenic factor across various cancer subtypes. Strong correlation between elevated IL8 levels and bladder cancer made it recognised as a potential biomarker for the disease (\cite{il8_2}), along with VEGF and PAI-1 \cite{novel_biom}.
    \item \textbf{IL-13 (urine)}: IL-13 is a cytokine acting as an immunomodulator. It is an important mediator in responses to parasitic infections, as well as in inflammatory processes, allergies, and cancers \cite{il13}. The link between bladder cancer and increased levels of IL-13 in blood was already established by Szymanska et. al \cite{il13_2}. 
    \item \textbf{PAI-1/tPA complex (serum)}: PAI-1 is an inhibitor of tPA (tissue plasminogen activator). Together they form a complex which regulates fibrin clot degradation through hydrolysis. High concentrations of PAI-1 increase the risk of blood clots, while low concentrations of PAI-1 are linked with an elevated risk of haemorrhages. Both tPA and PAI-1 proteins are recognised as factors involved in carcinogenesis \cite{tpa1,tpa1_2}. Elevated PAI-1 levels have been associated with various types of cancers, as the protein can protect tumour cells from apoptosis and confer resistance to therapeutic treatments \cite{pai,pai2}. In breast cancer, the prognostic value of the PAI-1/tPA complex has been shown to be comparable to that of total PAI-1, and it serves as a predictor of poor clinical outcome \cite{pai}. What is the most important in the context of the HaBio cohort? PAI-1 was identified as a potential biomarker for detecting bladder cancer (\cite{pai3,pai4}), especially when measured with IL8, VEGF and APOE in a panel \cite{novel_biom}. 
\end{enumerate}

Most of the 10 most relevant features shown in the heatmaps (Fig. 5-6, Fig. S1) are associated with bladder cancer or with carcinogenesis more broadly. The detection of many of them in urine samples strongly suggests that malignant processes are occurring within the urinary tract.
What is also notable is that there are sex-specific differences (only two of the top 10 biomarkers are common to both sexes), not only between features but also in threshold values for biomarkers that overlap across datasets, thereby confirming the previously proposed hypothesis that biomarkers characteristic of bladder cancer development differ by sex.
Hence, the results can be treated as a basis for a screening panel with sex-specific biomarkers for bladder cancer. Instead of evaluating individual features separately, it would be more informative to interpret them as a set of diagnostic rules. Because the established biomarker for bladder cancer, BTA, may be influenced by haematuria (\cite{bta3}), a panel of additional biomarkers would be desirable to confirm or exclude the presence of bladder cancer. From a clinical perspective, taking urinary samples rather than blood is more comfortable for patients. In addition, screening using urine samples does not require medical staff to draw blood from patients, thereby reducing the cost of this type of medical examination and the burden on the healthcare system.  
While biomarkers are important factors to assess the risk of BC for each patient, it is also important to take a closer look at the history of smoking, which translates to harmful tar exposure. It is the only feature that is not a biomarker and ranks highly. The risk for BC associated with tobacco smoking is established as ground truth and cannot be overlooked. In the male group, a potential confounding factor is comorbidity, as patients may present both with bladder cancer and prostate issues (e.g. BPE) at once, which could affect the classification results. Therefore, analysing a set of biomarkers measured at the same time point provides a more reliable basis for decision-making than relying on a single biomarker (e.g., BTA).

\newpage

\section*{DATA AND METHODS}

\subsection*{Data collection and curation}
Rather than treating missing values solely as a nuisance, we explicitly model increasing levels of missingness as a controlled stress test to assess model robustness and feature stability under realistic data degradation scenarios.
First, the data were cleaned, and EDA (exploratory data analysis) was performed on the HaBio dataset to prepare it for analysis. The HaBio cohort dataset includes 675 patients with haematuria, categorised into two groups: control and bladder cancer. To qualify as a control patient in the HaBio cohort study, individuals had to meet specific criteria, including no prior history of cancer, positive haematuria (past or present), a negative cystoscopy within the last 3 months, and no prior chemotherapy or radiotherapy.
Additionally, control patients were selected to match bladder cancer patients already enrolled in the study in terms of sex, approximate age range, and smoking status. On the other hand, patients in the bladder cancer group were required to have a history of positive haematuria (past or present) and have undergone cystoscopy within the last six months. They also could not have a history of cancer other than bladder cancer and needed to have either a confirmed bladder cancer diagnosis or a suspicion of the disease.  

\begin{figure}[h!]
    \centering
    \includegraphics[width=0.7\textwidth]{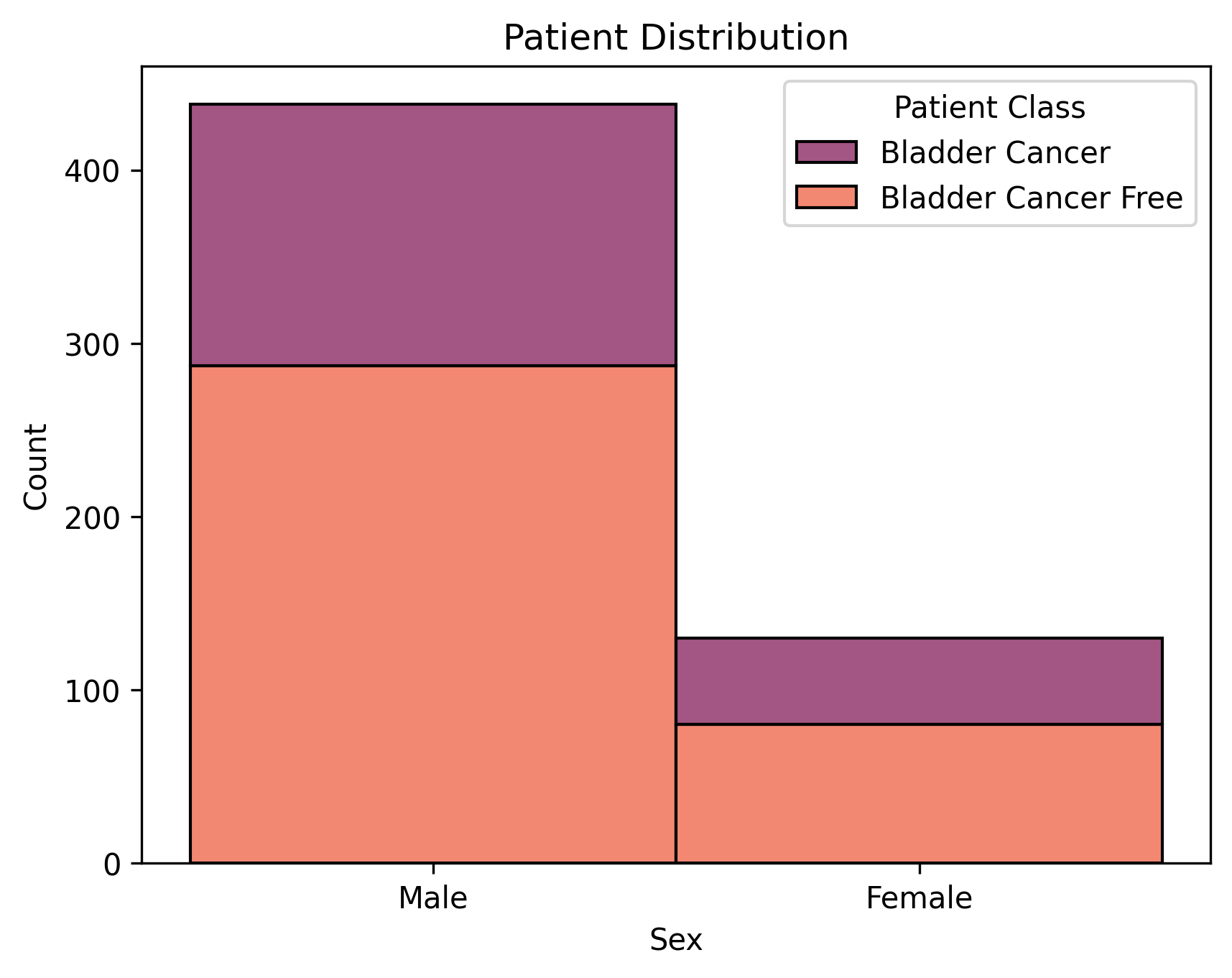}
    \caption{
    \textbf{Distribution of the HaBio cohort patients regarding sex}. 
    \newline{A bar plot illustrating the distribution of bladder cancer (BC) and bladder cancer free (non-BC) patients by sex.}
    \label{fig:distrib}
    }
\end{figure}

During the final diagnosis, 107 patients were labelled \textit{undiagnosed}, while 9 patients were initially misclassified as having bladder cancer but were later found to have an infection.
To eliminate this ambiguity, the 107 undiagnosed entries were removed, resulting in a final dataset of 568 patients, comprising 201 bladder cancer cases and 367 non-bladder cancer cases (including 9 entries initially classified as bladder cancer), which served as criterion/target class for classification. The dataset was notably unbalanced by gender, with 130 female patients (23\%) and 438 male patients (77\%) (Figure \ref{fig:distrib}). 

HaBio Cohort dataset includes multi-dimensional patient data collected at various stages of the study conducted between 17/10/2012 and 21/02/2020. At the time of recruitment, a nurse or HaBio clinician collected information on the patient’s age, sex, current medications, medical history, exposure risks, lifestyle choices, stimulant use, occupation, cystoscopy date, and bladder cancer diagnosis date, if applicable. It was also reported whether haematuria was visible with the naked eye. Urine and blood samples were analysed to quantify 79 biomarkers. During the follow-up period, the clinician conducted a detailed pathological review, including assessment of the conditions identified as the underlying causes of hematuria. 

Given that the original dataset contained a wide range of patient-related information, careful feature selection was necessary to retain only the most relevant variables for meaningful analysis. To prevent misleading conclusions caused by extreme data scarcity, features (columns) related to exposure data and medication details were excluded from the final dataset. The resulting dataset contained a well-balanced selection of features, providing a holistic view of each patient’s characteristics. It included demographic variables, lifestyle choices, general health information, occupational risk assessed by the Office of National Statistics and 79 biomarkers measured in the HaBio study. Subsequently, for a more detailed analysis, the dataset was stratified by sex, yielding 3 final cohorts: a combined dataset for both sexes (here named TOTAL), a subset of female patients (FEMALES), and a subset of male patients (MALES). 

The final dataset consisted of both continuous and categorical columns: “BlaCA\_noBlaCA” (target column storing whether the final diagnosis for a patient was diagnosed with bladder cancer (1) or not (0)), Haem\_Macro\_Micro (haematuria: macro or micro), ś80HdG (urine, ng/ml), Albumin - Creatinine Ratio ((ACR), urine), BTA (urine, U\/ml), CD44 (serum, ng/ml), CEA (serum, ng/ml), CK20 (urine, ng/ml), Clusterin (urine, ng/ml), Creatinine (urine, $\mu$mol/L), CRP (urine, mg/ml), CRP (serum, mg/ml), CXCL16 (urine, ng/ml), Cystatin-B (urine, ng/ml), Cystatin-C (urine, ng/ml), Cystatin-C (serum, ng/ml), D-dimer (urine, ng/ml), EGF (urine, pg/ml), EGF (serum, pg/ml), FABP-A (serum, ng/ml), FAS (urine, pg/ml), GRO (serum, pg/ml), HAD (serum, U/l), IFN-$\gamma$ (urine, pg/ml), IFN-$\gamma$ (serum, pg/ml), IL-1$\alpha$ (urine, pg/ml), IL-1$\alpha$ (serum, pg/ml), IL-1$\beta$ (urine, pg/ml), IL-1$\beta$ (serum, pg/ml), IL2 (urine, pg/ml), IL2 (serum, pg/ml), IL3 (urine, pg/ml), IL4 (urine, pg/ml), IL4 (serum, pg/ml), IL6 (urine, pg/ml), IL6 (serum, pg/ml), IL7 (urine, pg/ml), IL8 (urine, pg/ml), IL8 (serum, pg/ml), IL10 (urine, pg/ml), IL10 (serum, pg/ml), IL12p70 (urine, pg/ml), IL13 (urine, pg/ml), IL18 (urine, pg/ml), IL23 (urine, pg/ml), LASP-1 (serum, pg/ml), M30 (serum, U/l), M2PK (serum, ng/ml), MCP-1 (urine, pg/ml), MCP-1 (serum, pg/ml), Microalbumin (urine, mg/l), Midkine (urine, pg/ml), MMP9 (urine, ng/ml), MMP9/NGAL (urine, ng/ml), MMP9/TIMP (urine, ng/ml), NGAL (urine, ng/ml), NSE (urine, ng/ml),  osmolality (urine, mOsm), PAI-1/tPA (serum, ng/ml), pERK (urine, pg/ml), Progranulin (urine, ng/ml), Prolactin (serum, $\mu$lU/l), Protein (urine, mg/ml), PSA-TPSA (serum, ng/ml), S100A4 (serum, ng/ml), sIL2R$\alpha$ (serum, ng/ml), sIL6R (urine, ng/ml), TGF-$\beta$1 (urine, pg/ml), Thrombomodulin (urine, ng/ml), TNF$\alpha$ (urine, pg/ml), TNF$\alpha$ (serum, pg/ml), sTNFR1 (urine, ng/ml), sTNFR2 (urine, ng/ml), tPA (urine, ng/ml), VEGF (urine, pg/ml), VEGF (serum, pg/ml),  HDL (serum, $\mu$mol/L), LDL (serum, $\mu$mol/L), Triglycerides (serum, $\mu$mol/L) and Cholesterol (serum, $\mu$mol/L). The biomarker columns in the original dataset have a small proportion of missing values, ranging from 3 to 24 per cent, with a mean of 6 and a median of 3 across features.   
Additionally, datasets contained: sex and age columns, weekly alcohol intake (ml), daily fluid intake (ml), smoking status (past, current, never), smoking history (in years) and exposure to tar (mg), diabetes status, number of different medications taken daily, frequency of the urinary tract infections, and ONS ranking (assessment of workplace hazards). In total, the dataset comprises 89 features. 

\subsection*{CACTUS}
CACTUS (Comprehensive Abstraction and Classification Tool for Uncovering Structures) is based on a modified Naive Bayes classification algorithm \cite{cactus}. An innovative data processing component of CACTUS is an abstraction (discretisation) module that transforms feature values into 2 categories (Up and Down) using the ROC (Receiver Operating Characteristic) curve. The module enables effective interpretability of the results and ensures anonymisation. Following abstraction, classification is performed by matching each record (patient’s features) to each class and assigning the class that is most similar.  Each feature (column) is ranked by its discriminative strength, thereby highlighting its contribution to the classification process. Features with similar distributions across target classes (e.g., healthy vs. sick) are considered less informative, whereas those with distinct patterns across classes are more informative, as they better differentiate the classes. Feature ranking provides clinicians with interpretable insights into the model’s reasoning by identifying features that enable the record to be classified into the appropriate class, and it helps recognise biases and faulty behaviour, thereby improving the model and increasing trust in its outputs. 

\subsection*{CLASSIC EXPLAINABLE MACHINE LEARNING APPROACHES}
\begin{enumerate}
    \item \textbf{{Random Forest (RF)}} is a classification method based on a decision tree ensemble \cite{rf}. It creates multiple random subsets of the data, passes them to multiple decision trees, and then combines their outputs by majority voting. In majority voting, the final class label for a given sample is determined by the class most frequently predicted by the individual trees. Extracting the most important features from the RF model involves measuring the reduction in Gini impurity when a particular feature is used for splitting; the greater the impurity reduction, the more important the feature is to the model. As a result of feature importance calculations, one can obtain the features that are most informative for the classification task \cite{rf2}. 
    \item \textbf{Selected Gradient Boosting Methods (AdaBoost, XGB, LGBM, CatBoost)}: Gradient boosting (GB) is an ML approach that, similarly to RF, is based on an ensemble of decision trees. However, in GB, weak learners are trained sequentially to minimise the errors made by their predecessors, yielding a strong learner that makes accurate predictions of the target variable. \textbf{AdaBoost} was the first practical application of the GB algorithm that uses ensembles of decision trees \cite{ada1,ada2}. It assigns higher weights to misclassified samples, making them more relevant for subsequent trees. However, AdaBoost is not robust with large datasets and does not support missing values. To apply this algorithm to datasets with missing values, an imputation algorithm is required to impute the missing values with the most likely numerical value. In this study, mean imputation was applied, replacing missing values with the mean of non-missing values for each feature. Feature importance in the AdaBoost method is computed by first obtaining impurity-based importance values from each weak learner and then averaging them using the learners’ weights, which reflect their predictive performance. \textbf{EXtreme Gradient Boosting (XGB)} is another gradient boosting algorithm, designed to handle sparse datasets \cite{xgb}. Its prediction is computed as the sum of the outputs from all trees once the stopping criterion (e.g., a sufficiently small error) is met. Unlike AdaBoost, XGB supports missing values through its Sparsity Aware Split Finding algorithm. Feature importance in XGB is derived similarly to random forests, but instead of using Gini impurity, it is based on the average reduction in the model’s overall classification loss. Both methods quantify how much a feature helps reduce the loss, but in random forests, this is measured locally via impurity reduction at each split, whereas in XGB, it reflects the contribution to the overall loss reduction (global model improvement). \textbf{Light Gradient Boosting Machine (LGBM)}, similarly to XGB, is an alternative GB method based on decision trees ensemble \cite{lgbm}. It is optimised for large, high-dimensional datasets and can handle missing values natively. It can serve as an alternative to XGB, as feature importances can be calculated in the same way. Lastly, \textbf{CatBoost}, also a gradient boosting algorithm \cite{catb}. It was developed to efficiently handle categorical features, which are often problematic for other GB methods. In addition, it supports sparse datasets (missing values) and provides fast execution. In CB, feature importance is computed to quantify the increase in the overall loss function when a feature is excluded. Compared to XGB and LGBM, which report gain as an averaged loss reduction per split in the decision tree, CB returns the total (raw) loss reduction contributed by each feature. 
    
\end{enumerate}


\newpage

\section*{CONCLUSIONS}
We performed experiments using the HaBio dataset to classify patient cases into bladder cancer and non-bladder cancer groups. CACTUS achieved the best overall performance in terms of balanced accuracy and sensitivity across all 3 subgroups (total, males, and females) compared with classical ML methods. 
What stands out is CACTUS showing superior feature stability, which is important in the case of applying ML models to medical scenarios: the most important features for the model remained consistent across datasets with varying proportions of missing data and showed the highest degree of overlap even when missingness reached 30\%, confirming the robustness of the method.
Although bladder cancer serves as a concrete and clinically relevant case study, the primary contribution of this work lies in its data-centric perspective on trustworthy machine learning. The challenges addressed here—missing data, limited sample sizes, and the need for interpretable and stable feature selection—are common across biomedical and real-world datasets.
These findings are particularly relevant from a clinical perspective, as medical datasets are often incomplete. Moreover, we have demonstrated that the set of relevant biomarkers yields greater information than any individual biomarker. On average, CACTUS demonstrated the best feature stability for males and females and ranked second after Random Forest for the total population. 
This difference between men and women may be attributed to distinct biomarker patterns, as the most important features (biomarkers) vary between these subgroups. Our results demonstrate that CACTUS is a promising candidate as a trustworthy AI system for medical decision-making, reducing uncertainty associated with complex datasets and enabling the identification of key patterns that could support the development of novel bladder cancer screening assays. CACTUS is effective and stable not only with complete data but also with many missing fields, and it offers the possibility of integrating multi-dimensional data of various types (continuous, categorical).
Our findings demonstrate that predictive performance alone is insufficient to characterise model reliability. Models with comparable accuracy can exhibit markedly different behaviour in terms of feature stability, with implications for reproducibility, interpretability, and downstream decision-making. Feature stability, therefore, provides complementary information that should be considered alongside traditional evaluation metrics.
Future work could apply our approach by dividing the datasets into biomarker and lifestyle/demographic subsets. This distinction could help clarify the relative contributions of each type of data, given that biomarkers appear to be more important for classification and may overshadow lifestyle/demographic factors, while also providing insight into which factors patients can control to reduce their risk of developing bladder cancer. Moreover, larger datasets should be used to confirm our initial findings based on the presented cohort. 
It is also important to note that the thresholds calculated by CACTUS to abstract features are not the same as those set by diagnostic companies. Thresholds represent the best value that separates two classes (here, BC and non-BC cases), and thus do not carry the same information as thresholds set by medical institutions for diagnosis, which result from a combination of features used in the model rather than from the influence of a single characteristic. 
Additionally, we should consider the differential impacts of each feature across populations. Lastly, another benefit of using CACTUS is that it does not require prior domain knowledge in the field of ML model building, conversely to when classic ML models are built (multiple hyperparameters need to be established before constructing the model), as a result, CACTUS not only provides more reliable results explainability, which is important in the healthcare sector, but is also easy to use by medical staff without the need for lengthy and costly trainings. 
From a data science standpoint, this work emphasises the importance of evaluating how models respond to data perturbations rather than static datasets alone. Incorporating stability-oriented analyses enables more reliable knowledge extraction and supports the development of trustworthy machine learning systems in domains where data imperfections are the norm rather than the exception.
All the arguments presented here emphasise that CACTUS can successfully serve as a trustworthy AI framework for medical applications. 
By prioritising interpretability and feature stability under incomplete data, CACTUS contributes a generalizable framework for robust pattern discovery in complex datasets. This perspective complements performance-driven evaluation and supports more trustworthy data-driven insights across application domains, and suggests that stability-aware evaluation should be considered a standard component of pattern discovery pipelines operating on imperfect real-world data.


\section*{RESOURCE AVAILABILITY}


\subsection*{Lead contact}


Requests for further information and resources should be directed to and will be fulfilled by the lead contact, Paulina Tworek (p.tworek@sanoscience.org) or Jose Sousa (j.sousa@sanoscience.org).

\subsection*{Data and code availability}


\begin{itemize}
    \item The description of HaBio Clinical Study is presented at: https://www.isrctn.com/ISRCTN25823942.
    \item Access to the CACTUS tool is provided upon request by Jose Sousa (j.sousa@sanoscience.org).
    \item Any additional information required to reanalyze the data reported in this paper is available from the lead contact upon request.    
\end{itemize}

\section*{ACKNOWLEDGMENTS}


This project has received funding from the European Union's Horizon 2020 research and innovation programme under grant agreement No 857533 and from the International Research Agendas Programme of the Foundation for Polish Science No MAB PLUS/2019/13. 
The publication was created within the project of the Minister of Science and Higher Education "Support for the activity of Centers of Excellence established in Poland under Horizon 2020" on the basis of the contract number MEiN/2023/DIR/3796.
This project has received funding from the European Union’s Horizon 2020 research and innovation programme under grant agreement No 857524.

\section*{AUTHOR CONTRIBUTIONS}


Conceptualization, S.A.O. and P.T.; methodology, S.A.O. and P.T.; investigation, S.A.O. and P.T.; writing-–original draft, S.A.O., P.T., L.G., M.W.R., M.J.K., P.F. and J.S.; funding acquisition, J.S. and M.W.R..; resources,  S.A.O., P.T., L.G., M.W.R., M.J.K., P.F. and J.S.; supervision, J.S. and M.W.R.

\section*{DECLARATION OF INTERESTS}

Mark W. Ruddock, Mary Jo Kurt and Peter Fitzgerald are employees of Randox Laboratories Ltd.


\section*{DECLARATION OF GENERATIVE AI AND AI-ASSISTED TECHNOLOGIES}


During the preparation of this work, the authors didn't use any AI-assisted technologies, the authors reviewed and edited the content as needed and take full responsibility for the content of the publication.

\section*{SUPPLEMENTAL INFORMATION INDEX}




\begin{description}
  \item Figure S1-S4 and Tables S1-S12 in a Supplementary Materials PDF. 
\end{description}

\newpage



\newpage


\bibliography{references}

\bigskip


\newpage

\end{document}